\documentclass{article} 

\usepackage{amsmath,amsfonts,bm}

\def\eqref#1{equation~\ref{#1}}

\def\1{\bm{1}}

\DeclareMathAlphabet{\mathsfit}{\encodingdefault}{\sfdefault}{m}{sl}
\SetMathAlphabet{\mathsfit}{bold}{\encodingdefault}{\sfdefault}{bx}{n}

\DeclareMathOperator*{\argmax}{arg\,max}

\usepackage{booktabs}
\usepackage{caption}
\usepackage{placeins}
\usepackage{ntheorem}
\usepackage{siunitx}
\usepackage{tabularx}
\usepackage{changes}
\usepackage{tabu}
\usepackage{array,tabularx}
\usepackage{threeparttablex}
\usepackage{subfigure}
\usepackage{graphicx} 
\usepackage{url}
\usepackage{times}
\usepackage{hyperref}

\usepackage[accepted]{icml2019}

\icmltitlerunning{HyperGAN}

\begin{document}
\twocolumn[
\icmltitle{HyperGAN: A Generative Model for Diverse, Performant Neural Networks}

\begin{icmlauthorlist}
\icmlauthor{Neale Ratzlaff}{osu}
\icmlauthor{Li Fuxin}{osu}
\end{icmlauthorlist}
\icmlaffiliation{osu}{School of Electrical Engineering and Computer Science, Oregon State University}
\icmlcorrespondingauthor{Neale Ratzlaff}{ratzlafn@oregonstate.edu}
\icmlcorrespondingauthor{Li Fuxin}{lif@oregonstate.edu}
\vskip 0.3in

]

\printAffiliationsAndNotice{}
\newcommand{\fix}{\marginpar{FIX}}
\newcommand{\new}{\marginpar{NEW}}
\newcommand{\rpm}{\raisebox{.2ex}{$\scriptstyle\pm$}}
\newcommand\blfootnote[1]{%
  \begingroup
  \renewcommand\thefootnote{}\footnote{#1}%
  \addtocounter{footnote}{-1}%
  \endgroup
}

\begin{abstract}
Standard neural networks are often overconfident when presented with data outside the training distribution. 
We introduce HyperGAN, a new generative model for learning a distribution of neural network parameters.
HyperGAN does not require restrictive assumptions on priors, and networks sampled from it can be used to quickly create very large and diverse ensembles. 
HyperGAN employs a novel mixer to project prior samples to a latent space with correlated dimensions, and samples from the latent space are then used to generate weights for each layer of a deep neural network.
We show that HyperGAN can learn to generate parameters which label the MNIST and CIFAR-10 datasets with competitive performance to fully supervised learning, while learning a rich distribution of effective parameters.
We also show that HyperGAN can also provide better uncertainty estimates than standard ensembles by evaluating on out of distribution data as well as adversarial examples.
\end{abstract}

\section{Introduction}
It is well known that it is possible to train deep neural networks from different random initializations and obtain models that, albeit having quite different parameters, achieve similar loss values \citep{freeman-topology}. It has further been found that ensembles of deep networks that are trained in such a way have significant performance advantages over single models \citep{maclin-ensembles}, similar to the classical bagging approach in statistics. Ensembles are also more robust to outliers, and can provide uncertainty estimates over their inputs \citep{Lak-DeepEnsembles}. 

In Bayesian deep learning, there is a significant interest in learning approximate posterior distributions over network parameters. Past approaches mostly leverage variational inference to model this likely intractable distribution. \citep{Gal-MCdropout} formulated dropout as a Bayesian approximation, and showed that networks with dropout following each layer are equivalent to a deep Gaussian process \citep{Damianou:deepGPs13} marginalized over its covariance functions. They proposed MC dropout as a simple way to estimate model uncertainty. Applying dropout to every layer however, may result in underfitting of the data. Moreover, dropout only integrates over the space of models reachable from a single (random initialization).
As another interesting direction, hypernetworks \citep{DHypernetworks-Ha} are neural networks which output parameters for a target neural network. The hypernetwork and the target network together form a single model which is trained jointly. Originally, hypernetwork produced the target weights as a deterministic function of its own weights, but Bayesian Hypernetworks (BHNs) \citep{Krueger-BHN}, and Multiplicative Normalizing Flows (MNF) \citep{MNF} learn variational approximations by transforming samples from a Gaussian prior through a normalizing flow. Normalizing flows can model complicated posteriors, but they are composed of invertible bijections, which limits their scalability and the variety of learnable functions.

In this paper we explore an approach which generates all the parameters of a neural network in a single pass, without assuming any fixed noise models or functional form of the generating function. To keep our method scalable, we do not restrict ourselves to invertible functions as in flow-based approaches. We instead utilize ideas from generative adversarial networks (GANs). We are especially motivated by recent Wasserstein Auto-encoder \citep{Tolstikhin-WAE} approaches, which have demonstrated an impressive capability to model complicated, multimodal distributions. 

One of the issues in generating weights for every layer is the connectivity of the network. Namely, the output of the previous layer becomes the input of the next layer, hence the network weights must be correspondent in order to generate valid results. In our approach, we sample from a simple multi-dimensional Gaussian distribution, and propose to transform this sample into multiple different vectors. We call this procedure a \textit{mixer} since it introduces correlation to the otherwise independent random noise. Then each random vector is used to generate all the weights within one layer of a deep network. The generator is then trained with conventional maximum likelihood (classification/regression) on the weights that it generates, and an adversarial regularization keeps it from collapsing onto only one mode. In this way, it is possible to generate much larger networks than the dimensionality of the latent code, making our approach capable of generating all the weights of a deep network with a single GPU. 

Somewhat surprisingly, with just this approach we can already generate complete, multi-layer convolutional networks which do not require additional fine-tuning. We are able to easily sample many well-trained networks from the generator which each achieve low loss on the target dataset. Moreover, our diversity constraints result in models significantly more diverse than training with multiple random starts (ensembles) or past variational methods.

We believe our approach is widely applicable to a variety of tasks.
One area where populations of diverse networks show promise is in uncertainty estimation and anomaly detection. We show through a variety of experiments that populations of diverse networks sampled from our model are able to generate reasonable uncertainty estimates by calculating the entropy of the predictive distribution of sampled networks. Such uncertainty estimates allow us to detect out of distribution samples as well as adversarial examples. Our method is straightforward, as well as easy to train and sample from. We hope that we can inspire future work in the estimation of the manifold of neural networks.  

We summarize our contribution as follows:
\begin{itemize}
\vspace{-0.10in}
    \item We propose HyperGAN, a novel approach to approximating the posterior of neural network parameters for a target architecture. HyperGAN contains a novel mixer that mixes input noise into separate vectors that generate each layer of the network respectively.
    \vspace{-0.03in}
    \item Different from prior GANs, HyperGAN does not require repeated samples to start with (e.g. no need to train $1,000$ networks as a training set) but trains directly using maximum likelihood. This significantly improve training efficiency. The generated networks perform well without need for further fine-tuning.
    \vspace{-0.03in}
    \item On classification experiments, 100 network ensembles generated by HyperGAN significantly improves accuracy. To validate the uncertainty estimates given by ensembles from HyperGAN, we performed experiments on a synthetic regression task, an open-category classification task and an adversarial detection task.
    \vspace{-0.05in}
\end{itemize}

\section{Related Work}
The hypernetwork framework~\citep{DHypernetworks-Ha} introduced models where one network directly supervises the weight updates of another network.  Hypernetworks have also been used as implicit density estimators, to model more flexible variational approximations. \cite{pawlowski2017implicit} learn an approximate posterior by using a hypernetwork to generate weight samples from a Gaussian prior. The sampled weights are regularized such that they do not drift to far from the prior, using a kernel approximation to the KL divergence.  
The input to the hypernetwork is noted to be independent between layers, while we employ a mixer to learn complex correlations between generator inputs. 
In place of hypernetworks, \citep{MNF} used normalizing flows to model the effect of multiplicative auxiliary variables on a fully factorized Gaussian posterior over the weights. Normalizing flows are also used by \citep{Krueger-BHN}, who assume independent Gaussian weights, and predict scale and shift factors per unit.  

Generating parameters for neural networks is not strictly the purview of approximate Bayesian inference. Computer vision methods often use data driven approaches as seen in methods such as Spatial Transformer networks \citep{Jaderberg-STN}, or Dynamic Filter networks \citep{BrabandereDynFilter}. In these methods the filter parameters of the main network are conditioned on the input data, receiving contextual scale and shift updates from an auxiliary network. Our approach, however, generates weights for the whole network. Furthermore our predicted weights are highly nonlinear functions of the input, instead of simple affine transformations based on the input examples. 

GANs have also been used as a method for sampling the distribution captured by a Bayesian neural network (BNN) trained with Stochastic Gradient Langevin Dynamics (SGLD).  \citep{wangAPD2018} propose Adversarial Posterior Distillation (APD): instead of generating parameters from the BNN posterior using MCMC, a GAN trained on intermediate models during the SGLD process is used to sample from a BNN. However, the training samples from the SGLD training process are inevitably correlated, potentially reducing the diversity of generated networks. Our approach does not use correlated examples in training and hence it can generate more diverse networks. \citep{henning2018approximating} also use a GAN framework to predict the target weights directly. However, instead of using a discriminator to judge the predicted weights, a hypernetwork generates a set of target weights, and the discriminator observes tuples of $(image, label)$ which come from the dataset or are predicted by the generated weights. The objective of the hypernetwork is to generate parameters which classify the input data such that the discriminator in turn classifies the data-label tuples as coming from the training data. By contrast, our method does not depend on a discriminator to estimate the density ratio between real and generated outputs. Furthermore, our method learns a latent space of layer embeddings, which we use to achieve high variance among our predicted weights. 

Recently, \citep{Lak-DeepEnsembles} proposed Deep Ensembles, where adversarial training was applied to standard ensembles to smooth the predictive variance. However, adversarial training is an expensive training process. Adversarial examples are generated for each batch of data seen. We seek a method to learn a distribution over parameters which does not require adversarial training. 

Meta learning approaches use different kinds of weights to increase the generalization ability of neural networks. The first proposed method is fast weights \citep{hinton1987using} which uses an auxiliary network to produce weight changes in the target network, acting as a short term memory store. Meta Networks \citep{munkhdalai2017meta} build on this approach by using an external neural memory store in addition to multiple sets of fast and slow changing weights. \citep{ba2016using} augment recurrent networks with fast weights as a more biologically plausible memory system.
Unfortunately, the generation of each predicting network requires querying the base (slow) learner many times. Many of these methods, along with hyperparameter learning \citep{lorraine-stoc-hyperparam}, propose learning target weights which are deterministic functions of the training data. Our method instead captures a distribution over parameters, and provides a cheap way to directly sample full networks from a learned distribution.

\begin{figure}[ht!]
\vskip -0.15in
\centering
\includegraphics[width=.9\linewidth]{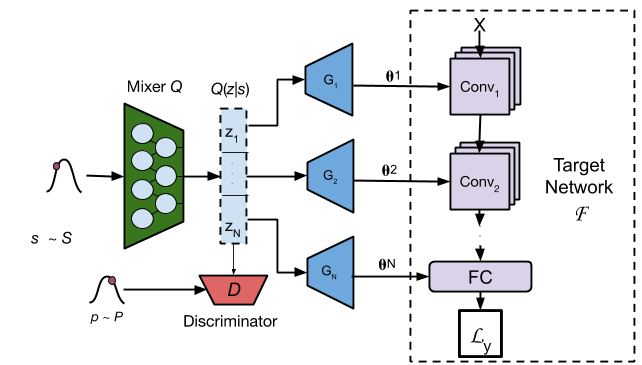}
\vskip -0.05in
\caption{HyperGAN architecture. The mixer transforms $s \sim \mathcal{S}$ into latent codes $\{z_1, \dots, z_N\}$. The generators transform each $q_i$ into parameters for the matching layer in the target network. The discriminator forces $Q(z|s)$ to be well-distributed and close to $\mathcal{P}$}
\label{fig:hypergan}
\vskip -0.1in
\end{figure}

\section{HyperGAN}
Taking note from the original hypernetwork framework \citep{DHypernetworks-Ha}, we coin our approach HyperGAN. The idea of HyperGAN is to utilize a GAN-type approach to directly model weights. To do this, a standard approach would be to acquire a large set of trained model parameters and use those as training data to the GAN~\cite{wangAPD2018}. 

However, a large collection of neural networks would be very costly to build. The other approach, proposed in ~\cite{wangAPD2018} to utilize many intermediate models during an SGLD training process as examples to train the GAN, results in too tightly correlated training samples. 

Instead, we propose to directly optimize the target supervised learning objective instead of focusing on reconstruction error on the training for the generator. Similar to a GAN, we start by drawing a random sample $s \sim \mathcal{S} = \mathcal{N}(\mathbf{0}, \mathbf{I}_j)$, where $0$ is an all-zero vector and $\mathbf{I}_j$ is a $j \times j$ identity matrix. The idea is that the random sample would provide enough diversity and if we can maintain such diversity while minimizing the supervised learning objective, we can generate models that all optimize the loss function well, but are sufficiently diverse as they are generated from different Gaussian random vectors.

Figure \ref{fig:hypergan} shows the HyperGAN architecture. We begin by defining a neural network as a function $\mathcal{F}(x;\theta)$ with input $x$ and parameters $\theta$, consisting of a given architecture with $N$ layers, and a training set $M$ with inputs and targets $(X,Y) = \{x_i,y_i\}_{i=1}^M$. 
Distinct from the standard GAN, we propose a \textit{Mixer} $Q$ which is a fully-connected network that maps $s \sim \mathcal{S}$ to a mixed latent space $Z \in \mathbb{R}^{Nd} , \,\, d < j$. The mixer is motivated by the observation that weight parameters between network layers must be strongly correlated as the output of one layer needs to be the input to the next one. Hence, it is likely that some correlations are also needed in the latent vectors that generate those weight parameters. Our $Nd$-dimensional mixed latent space $Q(z|s)$ contains vectors that are all correlated, which we then partition into $N$ layer embeddings $[z_1, \ldots, z_N]$, each being a $d$-dimensional vector. Finally, we use $N$ parallel generators $G = \{G_1(q_1) \dots G_N(q_N)\}$ to generate the parameters $\theta_G$ for all layers in $\mathcal{F}$. This approach is also memory efficient since the extremely high dimensional space of the weight parameters are now separately connected to multiple latent vectors, instead of fully-connected to the latent space.
We then evaluate the new model $\mathcal{F}(x, \theta_G)$ on the training set. We define an objective which minimizes the error of generated parameters with respect to task loss $\mathcal{L}$:
\begin{equation}
        \inf_{G, Q} \underset{s\sim \mathcal{S}}{\mathbb{E}} \mathbb{E}_{(x, y) \sim (X, Y)}\left[\mathcal{L}(\mathcal{F}(x; G(Q(s))), y)\right]
        \label{eq:prelim}
\end{equation}
At each training step we generate a different network $G(Q(s))$ from a random $s \sim S$, and evaluate the loss function on a mini-batch from the training set. 
The resulting loss is backpropagated through the generators until $\theta_G$ minimizes the target loss $\mathcal{L}$.  


The main concern about directly optimizing the formulation in (\ref{eq:prelim}) would be that the codes sampled from $Q(z|s)$ may collapse to the maximum likelihood estimate (MLE) (when $\mathcal{L}$ is a log-likelihood).
This means that the generators may learn a very narrow approximation of $\Theta$. On the other hand, one can think of the training process as simultaneously starting from many starting points (since we sample different $s$ for each mini-batch) and attempting to obtain a good optimum on all of them. Because deep networks are extremely overparameterized and many global optima exist \citep{choromanska2015loss}, the optimization may indeed converge to different optima from different $s$.

The mixer may make the training process easier by building in the required correlations into the latent code, hence improving the chance the optimization converges to different optima from different random $s$. To further ensure that the parameters are well distributed, we add an adversarial constraint on the mixed latent space $\mathcal{D}(Q(z|s))$ and require it to not deviate too much from a high entropy prior $\mathcal{P}$. This constraint is closer to the generated parameters and ensures that $Q(z|s)$ itself does not collapse to always outputting the same latent code. With this we arrive at the HyperGAN objective:
\begin{equation}
\begin{split}
   & \inf_{G, Q} \underset{s\sim \mathcal{S}}{\mathbb{E}} \mathbb{E}_{(x, y) \sim (X, Y)}\left[\mathcal{L}(\mathcal{F}(x; G(Q(s))), y)\right] \\
   & - \beta \mathcal{D}(Q(s), \mathcal{P})
    \label{eq:hypergan}
\end{split}
\end{equation}
Where $\beta$ is a hyperparameter, and $\mathcal{D}$ is the regularization term which penalizes the distance between the prior and the distribution of latent codes. In practice $\mathcal{D}$ could be any distance function between two distributions. We choose to parameterize $\mathcal{D}$ as a discriminator network $D$ that outputs probabilities, and use the adversarial loss \citep{goodfellow-GAN} to approximate $\mathcal{D}(Q(s), \mathcal{P})$. Note that in practice, $\mathcal{P}$ and $\mathcal{S}$ are multivariate Gaussians with different dimensionality and covariance. 
\begin{equation}
    \mathcal{D} := - \sum_{i=1}^N \left( \log D(p_i) + \log (1 - D(z_i))\right)
    \label{eq:Dz}
\end{equation}
Note that we find it difficult to learn a discriminator in the output (parameter) space because the dimensionality is high and there is no structure in those parameters to be utilized as in images (where CNNs can be trained). Our experiments show that regularizing in the latent space works well, which matches results from recent work in implicit generative models \citep{Tolstikhin-WAE}.

This framework is general and can be adapted to a variety of tasks and losses. In this work, we show that HyperGAN can operate in both classification and regression settings. For multi-class classification, the generators and mixer are trained with the cross entropy loss function:
\begin{equation}
\begin{split}
    &\quad \mathcal{L}_H = M^{-1} \sum_{i=1}^M \, y_i \log \big( \mathcal{F}(x_i; \theta) \big) \\
    &\quad \theta = \{G_1(z_1), \dots, G_n(z_n)\} 
\end{split}
\end{equation}
For regression tasks we replace the cross entropy term with the mean squared error (mse): 
\begin{equation}
\begin{split}
    & \mathcal{L}_{mse} = M^{-1}  \sum_{i=1}^M \, (y_i - \mathcal{F}(x_i; \theta))^2 \\  
    & \theta = \{G_1(z_1), \dots, G_n(z_n)\}
\end{split}
\end{equation}
\vskip -2em
\subsection{Discussion: Learning to Generate without Explicit Samples}
In generative models such as GAN or WAE \citep{Tolstikhin-WAE}, it is required to have a training set from the distribution that is being estimated, so that such training examples can be generated from the generator. HyperGAN does not have a given set of samples to train with. Instead, it optimizes a supervised learning objective such as maximum likelihood. To draw a connection between this objective and the traditional reconstruction objective in GAN, we note that after training, $\theta_G$ represents the maximum likelihood estimate of $\mathcal{F}(x; \theta)$, which has a well-known link to KL-divergence:
\begin{eqnarray}
\label{eq:klchange}
    & &\inf_{\theta_G} D_{KL}(P(x|\theta) || P(x| \theta_G)) \\ 
    & = &\inf_{\theta_G} \mathbb{E}_{P(x|\theta)} \left[ \log P(x| \theta) - \log P(x|\theta_G) \right] \nonumber\\
    & = &\inf_{\theta_G} \mathbb{E}_{P(x|\theta)} \left[ -\log P(x| \theta_G) \right] \nonumber
\end{eqnarray}
(\ref{eq:klchange}) shows that by minimizing the error of the MLE on the log-likelihood, we are indeed minimizing the KL divergence between the unknown true parameter distribution $\Theta$ and the generated samples $\theta_G \in \Theta_G$. This is a little different from the commonly used ELBO objective in Bayesian neural networks in that the ELBO would be minimizing $D_{KL}(P(x|\theta_G), P(x|\theta))$ and require sampling from the posterior, whereas here we only require the training set itself. 



Hence, we can view HyperGAN also as approximating a target distribution of neural network parameters. However, HyperGAN only assumes the target distribution exists, and update our approximation $\Theta_G$ via maximum likelihood to better match the unknown $\Theta$. 

\section{Experiments}
\subsection{Experiment Setup}
We conduct a variety of experiments to test HyperGAN's ability to achieve both high accuracy and obtain accurate uncertainty estimates. First we show classification performance on both MNIST and CIFAR-10 datasets. Next we examine HyperGAN's capability to learn the variance of a simple 1D dataset. Afterwards, we perform experiments on anomaly detection by testing HyperGAN on out-of-distribution examples. For models trained on MNIST we test on notMNIST. For CIFAR experiments we train on the first 5 classes of CIFAR-10 (airplane, automobile, bird, cat, deer), and consider the 5 remaining classes as out-of-distribution data which we test on. Finally, we test our robustness to adversarial examples as extreme cases of out-of-distribution data. 

In the following experiments we compare against APD \citep{wangAPD2018}, MNF \cite{MNF}, and MC Dropout \citep{Gal-MCdropout}. We also evaluate standard ensembles as a baseline. The target architecture used is the same across all approaches. More details about the target architecture can be found in the supplementary material.
For APD we train both MNIST and CIFAR networks with SGLD for 100 epochs. We then train a GAN given the architectures and hyperparameters specified in \citep{wangAPD2018} until convergence. For MNF we use the code provided by \citep{MNF}, and we train the model for 100 epochs. MC dropout is trained and sampled from as described in \citep{Gal-MCdropout}, with a dropout rate of $\pi = 0.5$. In all experiments, unless otherwise stated, we draw 100 networks from the posterior to form the predictive distribution for each approaches. For standard ensembles, we train both MNIST and CIFAR networks for 100 epochs, using Adam with a learning rate of $0.01$ which we decay as the loss plateaus. 

\subsection*{HyperGAN Details}
Both of our models take samples $s \sim S \in \mathbb{R}^{256}$ as input, but have different sized mixed latent spaces. For MNIST experiments, HyperGAN has weight generators, each taking a latent vector $z \sim Q(z|s) \in \mathbb{R}^{128}$ as input. The target network for the MNIST experiments is a small two layer convolutional network followed by 1 fully-connected layer, using leaky ReLU activations and 2x2 max pooling after each convolutional layer. For CIFAR-10, we use 5 weight generators with latent codes $z \sim Q(z|s) \in \mathbb{R}^{256}$. The target architecture for CIFAR-10 consists of three convolutional layers, each followed by leaky ReLU and 2x2 max pooling, followed by 2 fully connected layers.

The mixer, generators, and discriminator are each 2 layer MLPs with 512 units in each layer and ReLU nonlinearity. We found that larger generators offered little performance benefit, and ultimately hurt scalability.
We trained our HyperGAN on MNIST using less than 1.5GB of memory on a single GPU, while CIFAR-10 used just 4GB, making HyperGAN surprisingly scalable.

\begin{table*}[ht!]
    \centering
    \begin{tabular}{lccc|ccc|ccc}
        \toprule
        &\multicolumn{2}{c}{HyperGAN} & \multicolumn{1}{c} {} &\multicolumn{3}{|c|}{Standard Training} 
        &\multicolumn{3}{|c}{APD} \\
        \midrule
        & Conv1  & Conv2 & Linear & Conv1 & Conv2 & Linear & Conv1 & Conv2 & Linear\\
        \midrule
        Mean       & 7.49 & 51.10 & 22.01 & 27.05 & 160.51 & 5.97 & 2.63 & 5.01 & 17.4\\
        $\sigma$ & 1.59 & 10.62 & 6.01  & 0.31  & 0.51   & 0.06 & 0.22 & 0.41 & 1.43\\
        \bottomrule
    \end{tabular}
    \caption{$L_2$-norm statistics on the layers of ensembles sampled from HyperGAN, compared to standard networks trained from different random initializations as well as samples from the posterior learned by APD. All models were trained on MNIST to 98\% accuracy. Its easy to see that HyperGAN generates far more diverse networks}
    \label{2norm}
\end{table*}%

\subsection{Classification Accuracy and Diversity}
First we evaluate the classification accuracy of HyperGAN on MNIST and CIFAR-10.
Classification serves as an entrance exam into our other experiments, as the distribution we want to learn is over parameters which can effectively solve the classification task. 
We test with both single network samples, and ensembles. For our ensembles we average predictions from $N$ sampled models with the scoring rule $p(y | x) = \frac{1}{N} \sum_{n=1}^N p_n(y \, | \, x, \theta_n)$,
the results are shown in Table \ref{classification-resutls}. 
We generate ensembles of different sizes and compare against APD \citep{wangAPD2018}, MNF \citep{MNF}, MC dropout \citep{Gal-MCdropout}, as well as a baseline single network denoted as random start. 
For each method we draw 100 samples from the posterior to generate a predictive distribution and use the above scoring rule to compute the classification score. 

\begin{table*}[t]
    \centering
    \begin{tabular}{lcccc}
        \toprule
        Method        & MNIST         & MNIST 5000 & CIFAR-5        & CIFAR-10\\
        \midrule
        1 network     & 98.64 \rpm .3  & 96.69 \rpm .3  & 84.50 \rpm .6 & 76.32   \rpm .3 \\
        5 networks    & 98.75 \rpm .3  & 97.24 \rpm .14 & 85.51 \rpm .2 & 76.84 \rpm .1  \\
        10 networks   & 99.22 \rpm .09 & 97.33 \rpm .1  & 85.54 \rpm .2 & 77.52 \rpm .09 \\
        100 networks  & \textbf{99.31 \rpm .02} & \textbf{97.71 \rpm .05} & \textbf{85.81 \rpm .02} & \textbf{77.71 \rpm .03}\\
        APD           & 98.61          & 96.35          & 83.21         & 75.62\\
        MNF           & 99.30          & 97.52          & 84.00         & 76.71\\
        MC Dropout    & 98.73          & 95.58          & 84.00         & 72.75\\
        Random Start  & 99.14          & 97.09          & 83.84         & 74.79\\
        \bottomrule
    \end{tabular}
    \caption{Classification performance of each method on MNIST and CIFAR-10. CIFAR-5 refers to a dataset with only the first 5 classes of CIFAR-10. MNIST 5000 refers to training on only 5000 examples of MNIST, which has been used in prior work (e.g. ~\cite{Krueger-BHN}). When held to the same architecture, the ensemble from HyperGAN performs better than ensembles using other approaches (100 model ensembles are used for APD, MNF and MC Dropout)}
    \label{classification-resutls}
\end{table*}

In Table \ref{2norm} we show some statistics of the networks generated by HyperGAN on MNIST. We note that HyperGAN can generate very diverse networks, as the variance of network weights generated by the HyperGAN is significantly higher than standard training from different random initializations, as well as APD. More insights on the diversity of HyperGAN samples can be found in Section \ref{section:ablation}.

\subsection{1-D Toy Regression Task}
We next evaluate the capability of HyperGAN to fit a simple 1D function from noisy samples and generate reasonable uncertainty estimates on regions with few training samples. This dataset, first proposed by \citep{Hern-PBP}, consists of 20 training points drawn uniformly from the interval $[-4, 4]$. The targets are given by $y = x^3 + \epsilon$ where $\epsilon \sim \mathcal{N}(0, 3^2)$. We used the same target architecture as in \citep{Hern-PBP} and \citep{MNF}: a one layer neural network with 100 hidden units and ReLU nonlinearity trained with MSE. For HyperGAN we use two layer generators, and 128 hidden units across all networks. We used only a 64-dimensional latent space on this small task. 

\begin{figure}[!t]
\vskip -0.15in
\subfigure{\includegraphics[width=4.0cm]{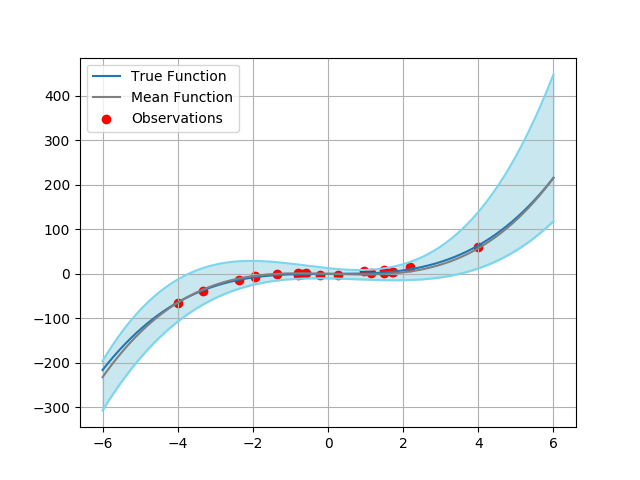}}
\subfigure{\includegraphics[width=4.0cm]{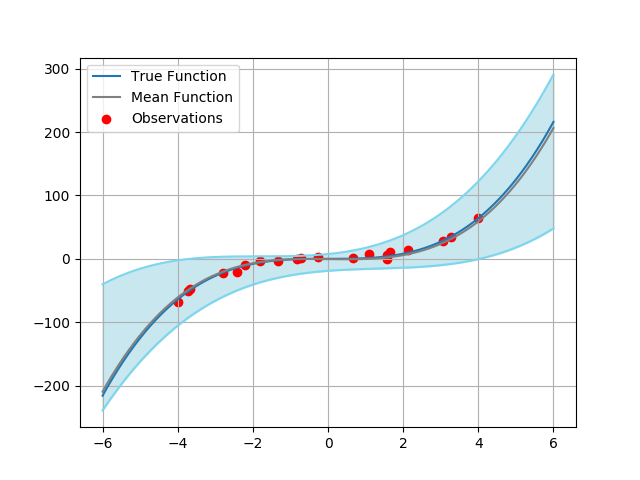}}
\vskip -0.15in
\caption{Results of HyperGAN on the 1D regression task. From left to right, we plot the predictive distribution of 10 and 100 sampled models from a trained HyperGAN. Within each image, the blue line is the target function $x^3$, the red circles show the noisy observations, the grey line is the learned mean function, and the light blue shaded region denotes $\pm3$ standard deviations}
\label{Regression}
\vskip -0.30in
\end{figure}

Figure \ref{Regression} shows that HyperGAN clearly learns the target function and captures the variation in the data. Furthermore, sampling more (100) networks to compose a larger ensemble improves the predicted uncertainty in regions with few training examples.

\subsection{Anomaly Detection}
To measure the uncertainty given on out of distribution data, we measure the total predictive entropy given by HyperGAN-generated ensembles. For MNIST experiments we train a HyperGAN on the MNIST dataset, and test on the notMNIST dataset: a 10-class set of 28x28 grayscale images depicting the letters A - J. In this setting, we want the softmax probabilities on inlier MNIST examples to have minimum entropy - a single large activation close to 1. On out-of-distribution data we want to have equal probability across predictions. Similarly, we test our CIFAR-10 model by training on the first 5 classes, and using the latter 5 classes as out of distribution examples. To build an estimate of the predictive entropy we sample multiple networks from HyperGAN, evaluate them on each example, and measure the entropy of the resulting distribution after the softmax. We compare our uncertainty (entropy) measurements with those of APD, MNF, MC dropout, and standard ensembles. Unless otherwise noted, we compute the entropy based on 100 networks.

\begin{figure}[ht!]
\vskip -0.15in
\centering
\includegraphics[width=.9\linewidth]{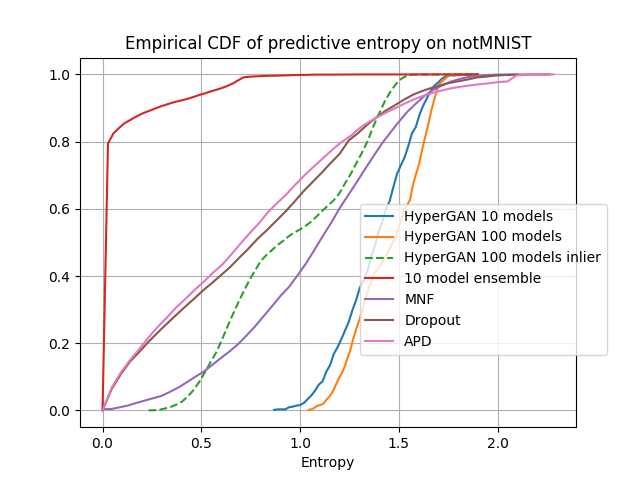}
\vskip -0.15in
\caption{Empirical CDF of the predictive entropy of all approaches on notMNIST. One can see the entropy of HyperGAN models are significantly higher than baselines}
\label{ANOMALY-DETECTION-MNIST}
\vskip -0.15in
\end{figure}

\begin{figure}[ht!]
\vskip -0.10in
\centering
\includegraphics[width=.9\linewidth]{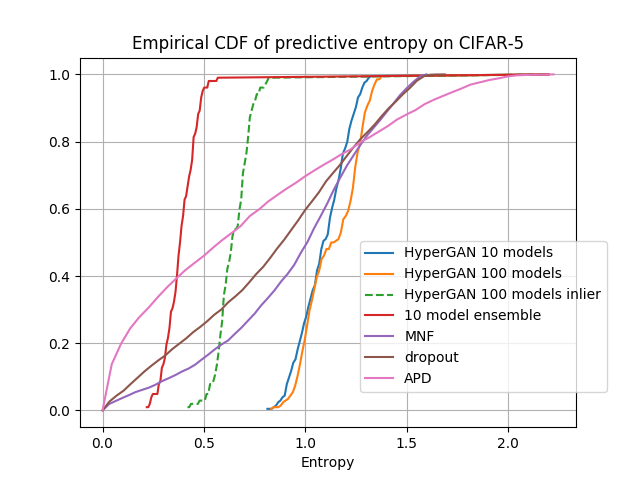}
\vskip -0.15in
\caption{Empirical CDF of the predictive entropy on out of distribution data: the 5 classes of CIFAR-10 unseen during training}
\label{ANOMALY-DETECTION-CIFAR}
\vskip -0.15in
\end{figure}

Fig. \ref{ANOMALY-DETECTION-MNIST} shows that HyperGAN is overall less confident on outlier samples than other approaches on the notMNIST dataset. Standard ensembles overfit considerably, as expected. Further, table \ref{2norm} shows that the diversity of standard ensembles is quite low.  Fig.  \ref{ANOMALY-DETECTION-CIFAR} shows similar behavior on CIFAR. Hence HyperGAN can better separate inliers from outliers when out-of-distribution examples are present.  

\subsection{Adversarial Detection}
We employ the same experimental setup to the detection of adversarial examples, an extreme type of out-of-distribution data. Adversarial examples are often optimized to lie within a small neighborhood of a real data point, so that it is hard for human to detect them. They are created by adding perturbations in the direction of the greatest loss with respect to the parameters of the model. Because HyperGAN learns a distribution over parameters, it should be more robust to adversarial attacks. We generate adversarial examples using the Fast Gradient Sign method (FGSM) \citep{goodfellow-fgsm} and Projected Gradient Descent with random restarts (PGD) \citep{madry-adversarial}. FGSM adds a small perturbation $\epsilon$ to the target image in the direction of greatest loss. FGSM is known to underfit to the target model, hence it may transfer better across many similar models. In contrast, PGD takes many steps in the direction of greatest loss, producing a stronger adversarial example, at the risk of overfitting to a single set of parameters. This poses the following challenge: to detect attacks by FGSM and PGD, HyperGAN will need to generate diverse parameters to avoid both attacks. 

To detect adversarial examples, we first hypothesize that a single adversarial example will not fool the entire space of parameters learned by HyperGAN. If we then evaluate adversarial examples against many newly generated networks, then we should see a high entropy among predictions (softmax probabilities) for any individual class. 

Adversarial examples have been shown to successfully fool ensembles \citep{dong-boosting}, but with HyperGAN one can always generate significantly more models that can be added to the ensemble for the cost of one forward pass, making it difficult to attack. 
We compare the performance of HyperGAN with ensembles of $N \in \{5, 10\}$ models trained on MNIST with normal supervised training. We fuse their logits (unnormalized log probabilities) together as $l(x) = \sum_{n=1}^N l_n(x)$ where $l_n$ is the logits of the $n$th model. For HyperGAN we simply sample the generators to create as many models as we need, then we fuse their logits together. Specifically we test HyperGAN ensembles with $N \in \{5, 10, 100, 1000\}$ members each. Adversarial examples are generated by attacking the ensemble directly until the generated image completely fools the whole ensemble. For HyperGAN, we attack the \textbf{full} ensemble, but test with a new ensemble of equal size. For other methods we follow their reported method, attacking a single model, then testing with 100 posterior samples. Note is disadvantageous to HyperGAN as it has to overcome a stronger attacker which has knowledge of the model distribution. 

\begin{figure*}[t]
\centering
\subfigure{\includegraphics[width=.39\linewidth]{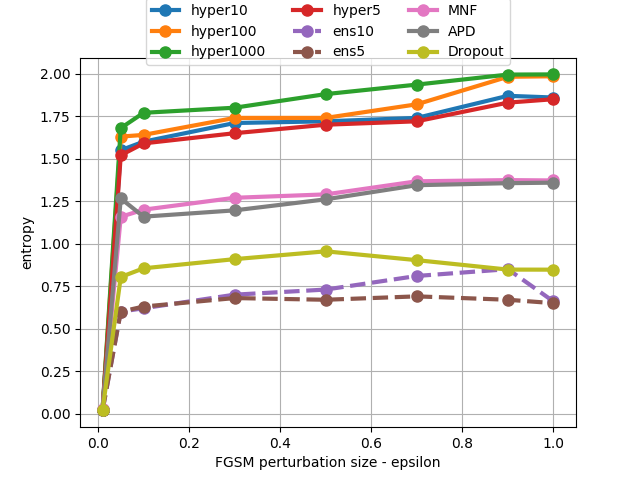}}
\hspace{.65in}
\subfigure{\includegraphics[width=.39\linewidth]{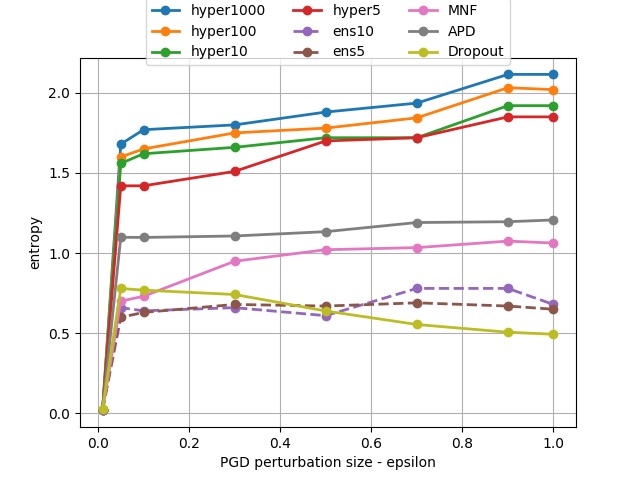}}
\vskip -0.2in
\caption{Entropy of predictions on FGSM and PGD adversarial examples. HyperGAN generates ensembles that are far more effective than standard ensembles even with equal population size. Note that for large ensembles, it is hard to find adversarial examples with small norms e.g. $\epsilon = 0.01$}
\label{adversarials}
\end{figure*}

For the purposes of adversarial detection, we compute the entropy within the predictive distribution of the ensemble to score the example on the likelihood that it was drawn from the training distribution. Figure \ref{adversarials} shows that HyperGAN predictions on adversarial examples approach higher entropy than other methods as well as standard ensembles, even when our experiment conditions are significantly disadvantageous for HyperGAN. HyperGAN is especially suited to this task as adversarial examples are optimized against a set of parameters - parameters which HyperGAN can change. Because HyperGAN can generate very diverse models, it is difficult for an adversarial example to fool the entirety of the ever-changing ensemble generated by HyperGAN. In the supplementary material we show that adversarial examples which fool a single HyperGAN sample fool only 50\% to 70\% of a larger ensemble.

\section{Ablation Study} 
\label{section:ablation}
The proposed architecture for HyperGAN is motivated by two requirements. First, we want to generate parameters for a target architecture which can solve a task specified by the data and the loss function. 
Second, we want the generated networks to be diverse. We address each concern with a specific element of our architecture. First, the mixer introduces the necessary correlations between generated layers allowing us to fit the target function. Second, the prior matching step using an adversarially trained discriminator enforces that samples from the mixed latent space $z \sim Q(z|s)$ are well-distributed. In this section we test the validity and effect of these two components by removing each one respectively and check the classification accuracy and the diversity of HyperGAN after the removal.

\begin{figure}[ht!]
\vskip -0.15in
\centering
\includegraphics[width=.85\linewidth]{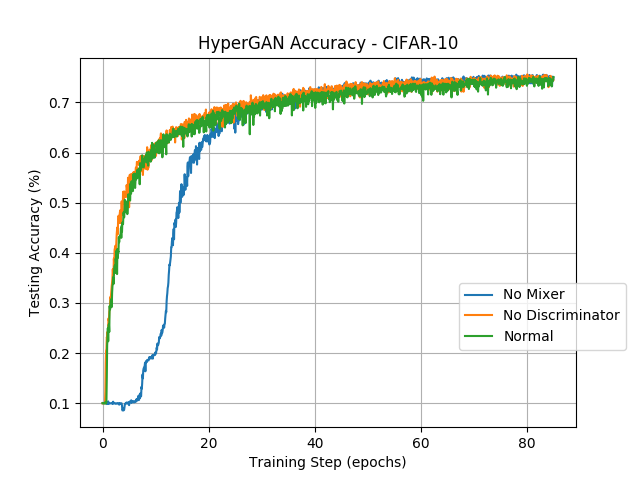}
\vskip -0.15in
\caption{Study of HyperGAN accuracy on CIFAR-10, with normal HyperGAN, without the mixer, and without the discriminator, respectively. Each test converges to similar accuracy but the variant without the mixer stumbles significantly in the beginning}
\label{ABLATION-ACC}
\vskip -0.05in
\end{figure}

\begin{figure}[ht!]
\vskip -0.15in
\centering
\includegraphics[width=.9\linewidth]{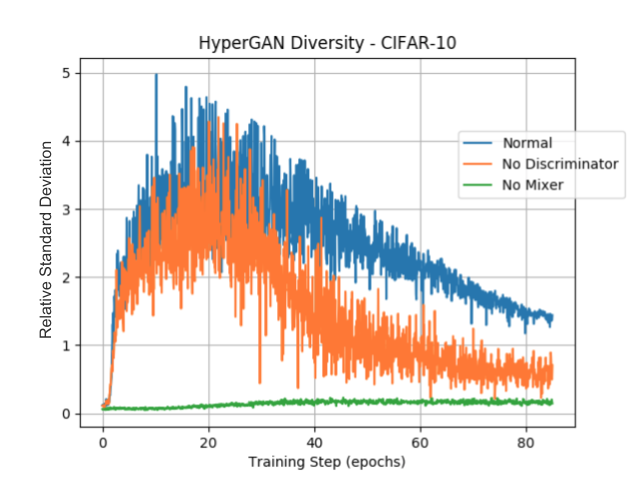}
\vskip -0.15in
\caption{HyperGAN diversity on CIFAR-10 given a normal training run, with the mixer removed, and with discriminator removed. Diversity is shown as the relative standard deviation of the $L_2$ norm of the weights, within a population of $100$ generated networks.}
\label{ABLATION-STD}
\vskip -0.35in
\end{figure}

First, we remove the regularization term $D(Q(s), \mathcal{P})$ from the training objective. Figure \ref{ABLATION-ACC} shows that the classification accuracy of the generated networks is unaffected. In figure \ref{ABLATION-STD}, we show diversity measured as the relative standard deviation between parameter sets sampled from HyperGAN during training. Since we can sample networks of similar accuracy, we want to be sure that the mixer is learning a wide distribution of parameters. Specifically, we measure the $L_2$ norm of $100$ weight samples and divide their standard deviation by the mean. We can see that without regularization, the diversity decreases. showing that by making the mixed latent space well-distributed, the discriminator is having a positive effect on the diversity of the generated models. This is similar to the prevention of mode collapse in adversarial ~\cite{Makhzani-AAE} and Wasserstein autoencoders~\cite{Tolstikhin-WAE}. We see that during training, diversity decreases over time, which is also common in GAN training. Hence, employing early stopping when accuracy has converged may be key to maintaining diversity as well. We would like to study the effect of early stopping more from the theoretical side in future work.

Next we remove the mixer $Q$ as well as the mixed latent space $Q(z|s)$. In this case, the generator for each layer takes as input an independent $d$-dimensional sample from an isotropic Gaussian. From Fig.~\ref{ABLATION-ACC}, we see that even in this case we can obtain similar classification accuracy. In Fig.~\ref{ABLATION-STD} we see that without the mixer, diversity suffers significantly. We hypothesize that without the mixer, a valid optimization trajectory is difficult to find (Fig.~\ref{ABLATION-ACC} shows that the HyperGAN with no mixer starts with low accuracy for a longer period); when one trajectory is finally found, the optimizer will prioritize classification loss over diversity. When the mixer is included, the built-in correlation between the parameters of different layers may have made optimization easier, hence diverse good optima are found even from different random starts.

\section{Conclusion and Future Work}
We have proposed HyperGAN, a generative model for learning approximate posteriors of neural network parameters. 
Training a GAN to learn a probability distribution over neural networks allows us to non-deterministically sample diverse, performant networks which we can use to form ensembles that obtain better classification accuracy and uncertainty estimates. Our method is scalable in terms of the number of networks in an ensemble, requiring just one forward pass to generate a new network, and a low memory footprint. We have also shown the uncertainty estimates from the generated ensembles are capable of detecting out-of-distribution data and adversarial examples. 

\section{Acknowledgements}
This work was partially supported by the Future for Life Institute grant 2017-174870 and DARPA contract N66001-17-2-4030. 
In addition we thank Dr. Alan Fern, Dr. Tom Dietterich, Lawrence Neal, Matthew Olson, and Alexander Turner for helpful discussions and proofreading. 

\bibliography{icml_bib}
\bibliographystyle{icml2019}

\appendix
\section{Appendix}
\subsection{Generated Filter Examples}
\begin{figure}[h!]
\centering
\subfigure[]{\includegraphics[width=.24\linewidth]{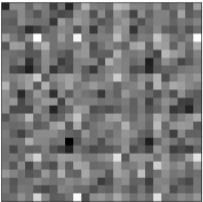}}
\subfigure[]{\includegraphics[width=.24\linewidth]{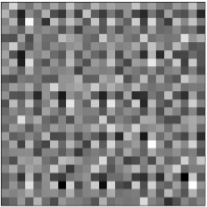}}
\subfigure[]{\includegraphics[width=.24\linewidth]{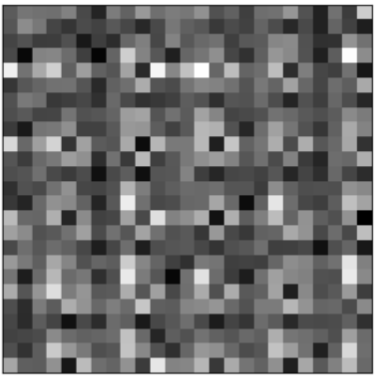}}
\subfigure[]{\includegraphics[width=.24\linewidth]{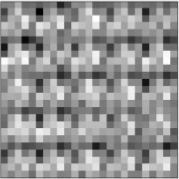}}

\caption{Convolutional filters from MNIST classifiers sampled from HyperGAN. For each image we sample the same 5x5 filter from 25 separate generated networks. From left to right: figures a and b show the first samples of the first two generated filters for layer 1 respectively. Figures c and d show samples of filters 1 and 2 for layer 2. We can see that qualitatively, HyperGAN learns to generate classifiers with a variety of filters.}
\label{fig:mnist_filters}
\end{figure}

We show the first filter in 25 different networks generated by the HyperGAN to illustrate their difference in Fig.~\ref{fig:mnist_filters}. It can be seen that qualitatively HyperGAN learns to generate classifiers with a variety of filters.

\subsection{Outlier Examples}

\begin{figure}[h!]
\centering
\subfigure[]{\includegraphics[width=\linewidth]{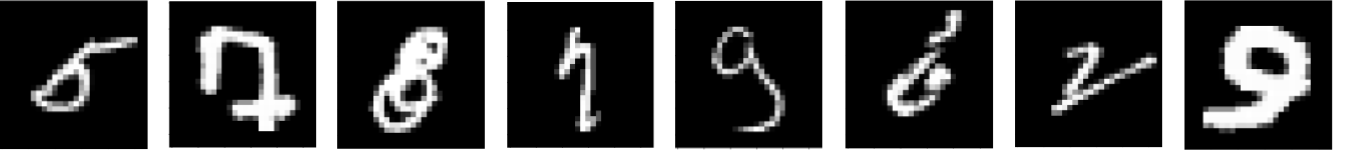}}\hspace{1em}%
\subfigure[]{\includegraphics[width=\linewidth]{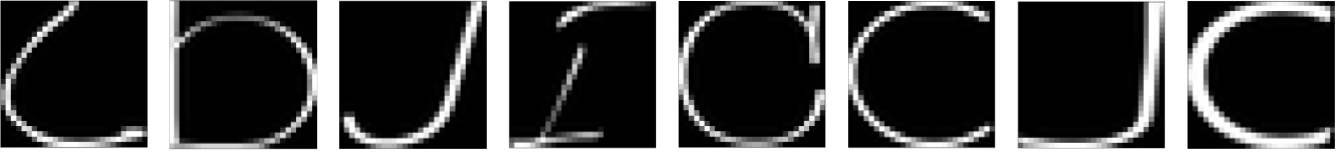}}\hspace{.9em}

\caption{Top: MNIST examples to which HyperGAN assigns high entropy (outlier). Bottom: Not-MNIST examples which are predicted with low entropy (inlier)}
\label{fig:mnist_samples}
\end{figure}

In Figure \ref{fig:mnist_samples} we show images of examples which do not behave like most of their respective distribution. On top are MNIST images which HyperGAN networks predict to have high entropy. We can see that they are generally ambiguous and do not fit with the rest of the training data. The bottom row shows notMNIST examples which score with low entropy according to HyperGAN. It can be seen that these examples look like they could come from the MNIST training distribution, making HyperGAN's predictions reasonable

\subsection{HyperGAN Network Details} \label{section:net_details}
In tables \ref{table:MNIST_SIZE} and \ref{table:CIFAR_SIZE} we show how the latent points are transformed through the generators to become a full layer of parameters. For a MNIST based HyperGAN we generate layers from small latent points of dimensionality $128$. For CIFAR-10 based HyperGANs we use a larger dimensionality of $256$ for the latent points.  
\begin{table}[htb]
\vskip -0.1in
    \centering
    \caption{MNIST HyperGAN Target Size}
    \label{table:MNIST_SIZE}
    \begin{tabular}{ccc}
        \toprule
        Layer & Latent size & Output Layer Size\\
        \midrule
        Conv 1    & 128 x 1 & 32 x 1 x 5 x 5\\
        Conv 2    & 128 x 1 & 32 x 32 x 5 x 5\\
        Linear    & 128 x 1 & 512 x 10\\
        \bottomrule
    \end{tabular}
    \caption{CIFAR-10 HyperGAN Target Size}
    \label{table:CIFAR_SIZE}
    \begin{tabular}{ccc}
        \toprule
        Layer & Latent Size & Output Layer Size\\
        \midrule
        Conv 1    & 256 x 1 & 16 x 3 x 3 x 3\\
        Conv 2    & 256 x 1 & 32 x 16 x 3 x 3\\
        Conv 3    & 256 x 1 & 32 x 64 x 3 x 3\\
        Linear 1  & 256 x 1 & 256 x 128\\
        Linear 2  & 256 x 1 & 128 x 10\\
        \bottomrule
    \end{tabular}
    \vskip -0.15in
\end{table}%

\subsection{Diversity with Neither Mixer nor Discriminator}
We run experiments on both MNIST and CIFAR-10 where we remove both the mixer and the discriminator. Tables \ref{MNIST-noMD} and \ref{CIFAR-noMD} show statistics of the networks generated by HyperGAN using only independent Gaussian samples to the generators. In this setting, HyperGAN learns to generate only a very small distribution of parameters.  

\begin{table}[ht]
    \centering
    \begin{tabular}[t]{cccccc}
        \toprule
        \multicolumn{6}{c}{HyperGAN w/o (Q, D) - CIFAR-10} \\
        \midrule
        & Conv1  & Conv2 & Conv3 & Linear1 & Linear2 \\
        \midrule
        Mean     & 1.87 & 16.83 & 9.35 & 10.66 & 20.35 \\
        $\sigma$ & 0.11 & 2.44 & 1.02 & 0.16 & 0.76  \\
        
        \toprule
        
        \multicolumn{6}{c}{Standard Training - CIFAR-10} \\
        \midrule
        & Conv1 & Conv2 & Conv3 & Linear1 & Linear2\\
        \midrule
        Mean & 5.13 & 15.19 & 16.15 & 11.79 & 2.45\\
        $\sigma$ &1.19 & 4.40  & 4.28  & 2.80  & 0.13\\
        \toprule
    \end{tabular}
    \caption{Statistics on the layers of networks sampled from HyperGAN without the mixing network or discriminator, compared to 10 standard networks trained from different random initializations
    }
    \label{CIFAR-noMD}
\end{table}%

\begin{table*}[!htbp]
    \vskip 1cm
    \centering
    \begin{tabular}[t]{lccc|ccc}
        \toprule
        &\multicolumn{2}{c}{HyperGAN w/o (Q, D) - MNIST} 
        &\multicolumn{1}{c} {} 
        &\multicolumn{3}{|c}{Standard Training - MNIST} \\
        \midrule
        & Conv1  & Conv2 & Linear & Conv1 & Conv2 & Linear\\
        \midrule
        Mean           &10.79 & 106.39 & 14.81  & 27.05 & 160.51 & 5.97\\
        $\sigma$     & 0.58 & 0.90 &  0.79  & 0.31 & 0.51 & 0.06\\
        \bottomrule
    \end{tabular}
    \caption{Statistics on the layers of a population of networks sampled from HyperGAN, compared to 10 standard networks trained from different random initializations. Without the mixing network or the discriminator, HyperGAN suffers from a lack of diversity}
    \label{MNIST-noMD}
\end{table*}%

\subsection{HyperGAN Diversity on Adversarial Examples}
\begin{figure}[ht!]
\hspace{1cm}
\vskip -.05in
\centering
\subfigure{\includegraphics[width=.49\linewidth]{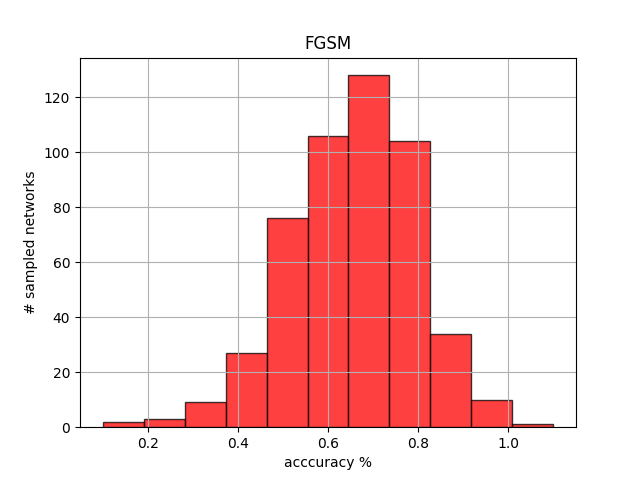}}
\hfill
\subfigure{\includegraphics[width=.49\linewidth]{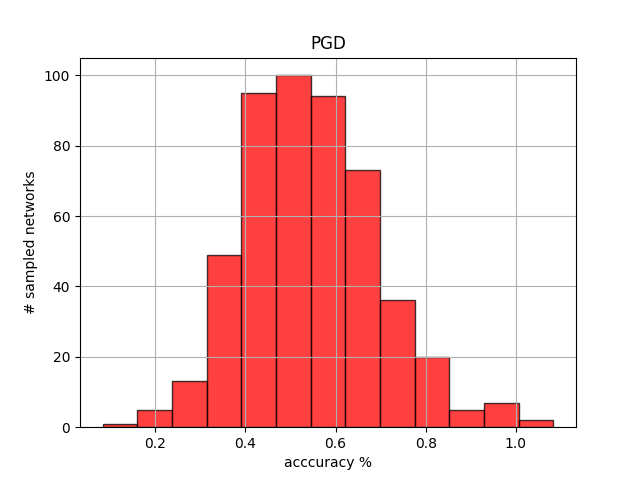}}
\hspace{1cm}
\vskip -0.2in
\caption{Diversity of predictions on adversarial examples. FGSM and PGD examples are created against a network generated by HyperGAN, and tested on 500 more generated networks. FGSM transfers better than PGD, though both attacks fail to cover the distribution learned by HyperGAN}
\vskip -0.1in
\label{diversity_adversarial}
\end{figure}

As an ablation study, in Fig.~\ref{diversity_adversarial} we show the diversity of the HyperGAN predictions against adversarial examples generated to fool one network. It is shown that while those examples can fool $50\% - 70 \%$ of the networks generated by HyperGAN, they usually never fool all of them.

\subsection{Black Box Adversarial Examples}
In addition to the white box attacks performed in section 4, we show here the results of HyperGAN on black box attacks. In the black box setting that we consider, the attack only has access to the $\argmax$ of the softmax probabilities. We test HyperGAN against a powerful decision attack called the Boundary Attack. The Boundary Attack begins from a large adversarial perturbation, then reduces the magnitude of the perturbation while retaining an adversarial prediction. We use the foolbox toolbox as before, with a step size of $0.01$ and a step multiplier of $1.5$. We report defense robustness in terms of number of calls (model evaluations) required by the attack to fool the classifier. We measure at [100, 500, 1000, 2000, 5000] calls, and see that HyperGAN performs similarly well as on white box attacks.

\begin{table}[!h]
\vskip -0.1in
    \centering
    \caption{HyperGAN performance on Black box attacks}
    \label{table:BB_attacks}
    \begin{tabular}{cccccc}
        \toprule
        Ensemble Size & 100 & 500 & 1000 & 2000 & 5000\\
        \midrule
        5 nets    & 1.85 & 1.79 & 1.75 & 1.73 & 1.55\\
        10 nets   & 1.88 & 1.79 & 1.76 & 1.76 & 1.63\\
        100 nets  & 1.92 & 1.80 & 1.75 & 1.76 & 1.70\\
        1000 nets & 1.98 & 1.81 & 1.78 & 1.75 & 1.72\\
        \bottomrule
    \end{tabular}
    \vskip -0.15in
\end{table}%

It is important to note that the size of the perturbation decreases with more calls. We therefore expect that robustness decreases as calls increase. 

\subsection{Effectiveness of Prior Matching}
In our work we encourage diversity by regularizing the mixer posterior $Q(z|s)$ to be well-distributed according to a high entropy distribution (a isotropic Gaussian in practice). We do so by sampling points from $Q(z|s)$ and estimate their distance from a prior $P$ with a discriminator $\mathcal{D}$. Regularizing the intermediate representation differs from standard variational inference, where samples from the posterior (generator outputs) are regularized to stay close to the prior. As stated in section 1, by regularizing the input samples, we retain flexibility in our generators to learn a more complex distribution over parameters than a unimodal high-dimensional Gaussian. To validate the effectiveness of the regularization we perform a normality test on $1000$ samples from $Q(z|s)$, drawn for each layer for a 3 layer target model.

\begin{table}[!h]
\vskip -0.1in
    \centering
    \caption{Normality test on points from $Q(z|s)$}
    \label{table:normality}
    \begin{tabular}{ccc}
        \toprule
        Layer & Mean & Std\\
        \midrule
        Conv 1    & 0.085 & 1.01\\
        Conv 2    & 0.014 & 1.20\\
        Linear    & 0.091 & 0.73\\
        \bottomrule
    \end{tabular}
    \vskip -0.15in
\end{table}%

We show the results from the normality test in table \ref{table:normality}. We can see that the induced distribution is approximately normal with means close to $0$. Furthermore, the mixer outputs distributions with different variance for each layer. The distribution with the highest variance corresponding to the second convolutional layer is what we expect, given that the representations of the final convolutional layer in a CNN are the most diverse. 

\end{document}